\documentclass[10pt,twocolumn,letterpaper]{article}

\usepackage{wacv}
\usepackage{times}
\usepackage{epsfig}
\usepackage{graphicx}
\usepackage{amsmath}
\usepackage{amssymb}

\usepackage{xcolor}
\usepackage{tabularx}
\usepackage{booktabs}
\usepackage{multirow}
\usepackage[numbers,sort,compress]{natbib}
\usepackage[utf8]{inputenc}
\usepackage{siunitx}

\wacvfinalcopy 
\ifwacvfinal
\def\assignedStartPage{1} 
\fi

\ifwacvfinal
\usepackage[breaklinks=true,bookmarks=falsem,colorlinks=true]{hyperref}
\else
\usepackage[pagebackref=true,breaklinks=true,colorlinks,bookmarks=false]{hyperref}
\fi

\ifwacvfinal
\else
\fi

\begin{document}

\title{Extraction of Positional Player Data from Broadcast Soccer Videos}

\author{Jonas Theiner$^1$\qquad Wolfgang Gritz$^1$ \qquad Eric Müller-Budack$^2$\\ \\ Robert Rein$^3$ \qquad Daniel Memmert$^3$ \qquad Ralph Ewerth$^{1,2}$\\ \\$^1$L3S Research Center, Leibniz University Hannover, Hannover, Germany \\ $^2$TIB -- Leibniz Information Centre for Science and Technology, Hannover, Germany \\$^3$Institute of Exercise and Sport Informatics, German Sport University Cologne \\{\tt\small $\{$theiner, ewerth$\}$@l3s.de} \qquad {\tt\small $\{$r.rein, d.memmert$\}$@dshs-koeln.de}
}

\maketitle
\ifwacvfinal\thispagestyle{empty}\fi
\ifwacvfinal\pagestyle{empty}\fi


\begin{abstract}
Computer-aided support and analysis are becoming increasingly important in the modern world of sports. The scouting of potential prospective players, performance as well as match analysis, and the monitoring of training programs rely more and more on data-driven technologies to ensure success. Therefore, many approaches require large amounts of data, which are, however, not easy to obtain in general. In this paper, we propose a pipeline for the fully-automated extraction of positional data from broadcast video recordings of soccer matches. In contrast to previous work, the system integrates all necessary sub-tasks like sports field registration, player detection, or team assignment that are crucial for player position estimation. The quality of the modules and the entire system is interdependent. A comprehensive experimental evaluation is presented for the individual modules as well as the entire pipeline to identify the influence of errors to subsequent modules and the overall result. In this context, we propose novel evaluation metrics to compare the output with ground-truth positional data. 
\end{abstract}

\section{Introduction}

Match analysis in soccer is very complex and many different factors can affect the outcome of a match. The question is which so-called key performance parameters allow for the characterization of successful teams~\cite{low2020, memmert2018data, Rein2016, sarmento2018}. 
While team behavior can be differentiated into a hierarchical scheme consisting of individual, group, and team tactics, different metrics are necessary to capture behavior at each level~\cite{garganta2009, Rein2016}. 
Researchers have recognized that game plays should be segmented into different phases since tactics vary greatly~\cite{mackenzie2013} across them. 
Performance in soccer is also determined by physiological factors~\cite{drust2007} such as running distance~\cite{DiSalvo2007}. For this reason, it has been suggested to link such information to tactical parameters~\cite{bradley2018}. 

\begin{figure*}[t]
\begin{center}
\includegraphics[width=1.0\linewidth]{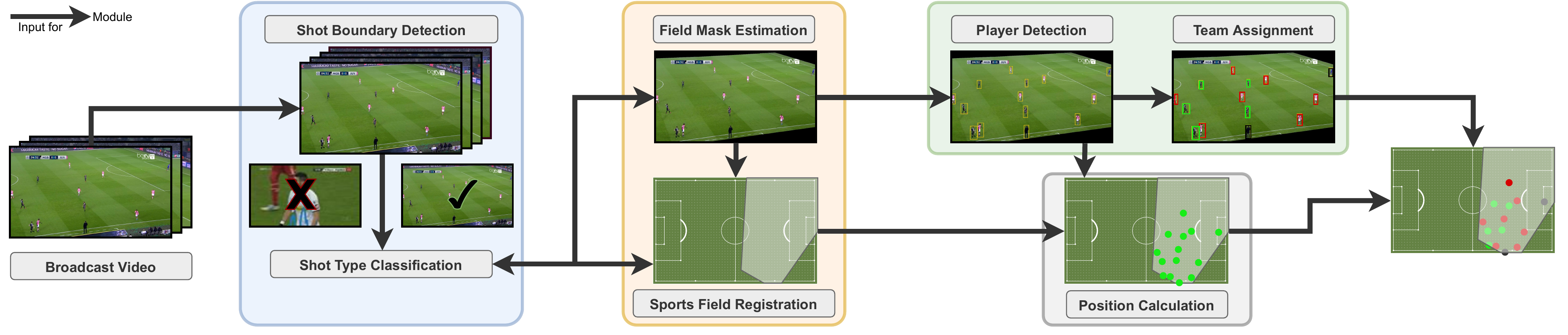}
\end{center}
   \caption{Proposed pipeline to extract positional data with team assignment from broadcast videos: The video is pre-processed to segment the field and detect shot boundaries. The camera type is estimated to extract shots from the main 
   camera. Subsequently, the sports field is registrated and the extracted homography matrix is used to transform the sport field and player detections in order to obtain two-dimensional coordinates for each player. Team assignment is performed by clustering the player's bounding boxes.}
\label{fig:pipeline}
\end{figure*}

To carry out such analyses, the player positions on the field are required.
Current tracking technologies allow the recording of several million data points representing player and ball positions during a match by using additional hardware, e.g., multiple static cameras or sensors on players.
However, they are difficult to obtain, for instance, due to licensing, financial restrictions, or competitive concerns, i.e., a club normally does not want or disclose its own team's data.
In contrast, broadcast video recordings of soccer matches can be accessed more easily.
In this paper, we introduce a 
modular pipeline to extract the two-dimensional positions of the visible players from ordinary broadcast recordings. 
As illustrated in Figure~\ref{fig:pipeline}, the system involves sports field registration
, shot boundary detection, shot type classification, player detection, and team assignment.

\textbf{Application Novelty:}
While commercial approaches like \cite{statsperform, tracab, metricasports} primarily use multiple static cameras for position data generation from video data, the TV feed is concretely used by 
\emph{SkillCorner}~\cite{skillcorner} and \emph{Track160}~\cite{track160}.
However, only the final output of such systems is accessible ~\cite{skillcorner, track160}. 
To the best of our knowledge, neither their quality, nor used architectures or even information about training data and applicability to own data is publicly reported.
While individual sub-tasks were tackled in research, its combination for the joint real-world task of \emph{player position estimation} 
has not been studied yet~(also not beyond soccer).
Even individual sub-modules have not been sufficiently evaluated in terms of applicability to real-world data.
For the essential step of sports field registration, recent approaches~\cite{sha2020end, nie2021robust} are evaluated only on a single small-scale dataset~\cite{homayounfar}. 
Potential for generalization were mentioned~\cite{nie2021robust, cioppa2021camera} with the use of many cost-intensive annotations from various data sources.
Furthermore, the influence of errors in individual modules and their connections has not been explored.
To tackle this demanding real-world task is of interest for the computer vision community as well for sports science, and has direct applications.


\textbf{Contributions:}
In contrast to commercial systems and related work, we provide the first transparent baseline for~\emph{player position estimation} with interchangeable modules, that relies on state-of-the-art techniques and freely available data, while evaluating each module.
We demonstrate the generalizability on multiple datasets where the applied models were not originally trained on.
The proposed pipeline is also applicable to the so-called "tactic-cam" that is located next to the main camera. It usually covers the entire soccer field~(without any cuts) and is consequently of interest for video analysts.
To evaluate the global task, estimated positions are compared to ground-truth positional data. 
This comparison is not trivial due to non-visible players in the video and the influence of errors of individual modules. 
Therefore, we propose novel evaluation metrics and identify the impact of errors on the final system output.

The remainder of the paper is organized as follows. Section~\ref{sec:rw} gives a brief overview of related work.
The pipeline itself is introduced in Section~\ref{sec:method}. In Section~\ref{sec:experiments}, the different system components and the accuracy of the extracted positional data are evaluated. Finally, Section~\ref{sec:conclusion} discusses the results and describes possible areas of future research.

\section{Related Work}
\label{sec:rw}

Since the global task of \textit{player position estimation} has not yet been addressed, we briefly review related work for all individual sub-tasks in this section.

Great progress has been made in recent years for \textbf{sports field registration} with monocular non-static cameras.
\citet{cuevas2020automatic} trained a probabilistic decision tree to classify specific line segments as an intermediate step for homography estimation and integrated a self-verification step to judge whether a predicted homography matrix is correct.
\citet{homayounfar} propose a solution that relies on field segmentation and Markov random fields. \citet{sharma} and \citet{chen2019} propose the nearest neighbor search from a synthetic dataset of pairs of edge images and camera images for fully-automated registration. 
\citet{error_refine} present a two-step deep learning approach that initially estimates a homography and minimizes the error using another deep network instead of the Lucas-Kanade algorithm~\cite{baker2004lucas}.
\citet{citraro2020} suggest an approach that also takes into account the position of players and is trained on a separate dataset for uncalibrated cameras. 
\citet{sha2020end} propose an end-to-end approach for area-based field segmentation, camera pose estimation, and online homography refinement that allows end-to-end training and efficient inference.
\citet{nie2021robust} tackle the challenge when no prior knowledge about the camera is available and propose a multi-task network to simultaneously detect a grid of key points and dense field features to estimate and refine a homography matrix end-to-end. This approach seems suitable since also temporal consistency is verified for successive frames.
However, a very large number of training samples is required to achieve the desired accuracy and generalizability, but training data are not publicly available except for the \textit{WorldCup2014} dataset~(\emph{WC14}~\cite{homayounfar}). 

\textbf{Shot boundary detection}~(e.g.,~\cite{tang2018fast,gygli2018ridiculously,wu2019two, transnet}) and \textbf{shot type classification}~(e.g.,~\cite{tong2015cnn, savardi2018shot}) are necessary pre-processing steps for many tasks of video analysis.
It enables the distinction between different camera shot types.
Related work in the context of soccer distinguishes between three~\cite{counterattack}, four~\cite{cnn_shot} or five~\cite{dataset_shots} different camera shot types.
For the extraction of positional data, the main camera~(with the largest distance) offers the most useful information, because it normally covers a larger part of the field depicting several players.

There are several approaches for the \textbf{detection of players} in sports analysis~\cite{FootAndBall2020, light_cascaded, player_detec_Direkoglu, ssd_playerdetect}.
Although general-purpose approaches for object detection~\cite{fasterrcnn, SSD} are also able to detect persons, sports offer specific challenges. For example, the players are usually small, they differ in scale due to the distance from the camera, they can occlude one another, and there is blur caused by camera movement. 
Nevertheless, specialized approaches~\cite{FootAndBall2020, ssd_playerdetect} compare themselves to general-purpose detectors such as the \emph{Single Shot Detector~(SSD)}~\cite{SSD} or \emph{Faster R-CNN}~\cite{fasterrcnn}. \citet{FootAndBall2020} have recently introduced a computationally much more efficient method with results similar to a fine-tuned \emph{Faster R-CNN}.
In team sports, the jerseys of the teams are designed so that they can be easily recognized by their color. Thus, for \textbf{team assignment} of the (detected) players, color information can be used as a discriminant feature.
Hand-crafted (color)~features~(\cite{d2009investigation, lu2013learning, tong2011automatic}) or features from convolutional neural networks~(CNNs)~(\cite{team_assign, lu2018lightweight, koshkina2021contrastive}) are exploited and clustered by these approaches for team assignment. 
An approach for player detection and team discrimination~\cite{team_assign} addresses the problem of occlusions and errors in object detection~\cite{manafifard2017survey}. 

\section{Player Position Estimation in Soccer Videos}
\label{sec:method}

A frequent problem in the field of automatic sports analysis is the lack of publicly available datasets. Currently, there is no public dataset that provides positional data for given broadcast soccer videos.
Besides, related work solely considered sub-problems of the overall task of \emph{player position estimation}. 
This section describes a pipeline as well as the choice and modifications of individual components that solve all required  sub-tasks for \emph{player position estimation} to predict the two-dimensional player positions on the field given an input (broadcast) video~(Figure~\ref{fig:pipeline}).

After all relevant (main camera) shots are identified~(Section~\ref{sec:exp:shots}), the step of sports field registration is essential to extract position data~(Section~\ref{sec:field_red}). 
A homography matrix is determined and used to transform the positions of the players from the image plane into world coordinates~(Section~\ref{sec:player_det_pos_est}).  

\subsection{Shot Boundary and Shot Type Detection}\label{sec:exp:shots}
We aim at estimating player positions in frames recorded by the main camera since it is most frequently used and shows the area of the game that is relevant for tactical analysis, as shown in Figure~\ref{fig:pipeline}. 
We first extract shots from the television~(TV) broadcast using the widely applied \emph{TransNet}~\cite{transnet, transnetv2} for shot boundary detection. Since our objective is to gather only valuable positional data, we subsequently apply shot type classification to identify shots captured by the main camera. 
We exploit the homography matrices estimated by the sports field registration approach presented in Section~\ref{sec:field_red}. We found that the homography matrices do not change fundamentally in successive frames captured by the main camera. On the other hand, all other cameras that, for example, capture player close-ups or actions depict no or only small fractions of the sports field causing large errors and consequently inconsistencies in the predicted homography matrices.
For this reason, we calculate the average $\overline{\mathcal{L}}_H$ of the homography changes for each shot. The homography change for two successive frames $t$ is defined as $\mathcal{L}_H(H_t, H_{t+1})=\Vert H_t - H_{t+1} \Vert_2$ where each entry in $H$ is (min-max) normalized for each shot.
Finally, we classify each shot as the main camera shot if the condition~$\overline{\mathcal{L}}_H \leq \tau$ is fulfilled. 

\subsection{Sports Field Registration}\label{sec:field_red}
The task of sports field registration aims at determining a homography matrix~$H$ for the transformation of an image from the (main) camera into two-dimensional sports field coordinates. 
Formally, the matrix~$H$ defines a two-dimensional projective transformation and is defined by a $3\times 3$ matrix with eight degrees of freedom. 
We use Chen and Little's approach~\cite{chen2019} as the basis for sports field registration. 
The camera calibration is defined as the nearest neighbor search in a synthetic dataset of edge map camera pairs. 
We choose this approach for multiple reasons: (1) It obtains almost state-of-the-art performance on the only test set for soccer~\cite{homayounfar}, (2) it does not rely on manual annotations to obtain training data~\cite{nie2021robust, error_refine, cioppa2021camera}, and (3) is adaptable to other environments~(e.g., stadiums and camera parameters) by changing only a few hyper-parameters, as shown in our experiments~(Section~\ref{exp:h_estimation}). 

\citet{chen2019} adopt a \emph{pix2pix}~\cite{isola2017image} model for field segmentation and the subsequent detection of the field markings.
The edge images generated in this way are compared with a dataset of synthetic edge images for which 
the camera parameters are known~($x, y, z$ position, focal length, pan, tilt). 
This comparison is based on a Siamese~CNN~\cite{hadsell2006dimensionality}, which takes two edge images as input. 
Feature vectors are used to construct the reference database.  
The nearest neighbor search on the feature vectors is then applied by computing the $L2$~distance over all pairs.
The camera parameters of the nearest neighbor in the synthetic dataset are used to determine an initial homography matrix.
This initial estimation is refined using the Lucas-Kanade algorithm~\cite{baker2004lucas}.

\subsection{Player Detection and Position Estimation}\label{sec:player_det_pos_est}
Sports analysis offers some specific challenges for the task of object detection and tracking, e.g., the objects (like players) are often small because they are far away from the camera. 
Camera motion causes blur in the players' silhouettes. But the movement with unpredictable changes of players' direction and pace poses problems also for well-tested approaches. 
Therefore, some approaches address these problems in the architectural design~\cite{FootAndBall2020, light_cascaded}.
\citet{centertrack} solves object detection and tracking based on the object center and should therefore be less susceptible to movements of the players.
In Section~\ref{exp:player_detection}, a comparison of three approaches is performed.


To determine the actual position of each player on the field, we can utilize the predicted homography matrix~$H$, which maps pixel coordinates to sports field coordinates. 
We define the image position~$\boldsymbol{\Tilde{p}} \in \mathbb{R}^2$ of players as the center of the bottom of the detected bounding box, which usually corresponds to the feet of the player. 
The predicted position $\boldsymbol{\hat p} \in \mathbb{R}^2$ of the player on the field is then calculated with the inverse homography matrix and the detected image positions of the players: $\boldsymbol{\hat p} = H^{-1} \boldsymbol{\Tilde{p}}$.

\textbf{Self-Verification (sv):}\label{self-verification}
The predicted positions can be used to verify the homography matrix extracted by the sports field registration. 
Assuming that most player positions should be assigned to a coordinate within the sports field, the system can automatically discard individual frames where the sports field registration is obviously erroneous. 
If one of the projected player positions 
is far outside the dimensions of the field including a tolerance distance~$\rho$ in meter, then normally there is an error in the homography estimation. 
The smaller the value~$\rho$ is chosen, the more frames are discarded, because only smaller errors in the homography estimation are being tolerated.
Intuitively, a tolerance distance between two and five meters seems reasonable which is proven experimentally~(Section~\ref{exp:position_estimation}).

\paragraph{Team Assignment:}
Assuming that for some sports analytic tasks the position of the goalkeeper is of minor relevance~(e.g., formation or movement analysis) and it is extremely rare that both goalkeepers are visible in the video at the same time, they are ignored in the team assignment step. 
Due to the different jersey type and color it requires context information (i.e. the location) to correctly assign the team.

Another problem is that coaches and attendants also protrude onto the sports field with their bodies due to the perspective of the camera so that the number of visible classes which appear in a frame cannot be predetermined. 
We present a simple approach that provides a differentiation between only two classes~(team A and B) based on the object detection and assumes that the use of an unsupervised clustering method is more appropriate in this domain since it does not rely on any training data and the player detection results are already available with high quality. 
We apply \emph{DBScan}~\cite{ester1996density} 
to determine two dominant clusters representing the field players of both teams.
Any unassigned detection, which should include goalkeepers, referees, and other persons, is discarded.
The feature vectors are formed based on the player detection results, i.e., the bounding boxes.
We use the upper half of a bounding box since it usually covers the torso of a player.
Each bounding box is first uniformly scaled to $20 \times 20$ and then the center of size $16 \times 16$ is cropped. 
This should reduce the influence of the surrounding grass in the considered area.
Since the jersey colors differ greatly, it is sufficient to use the average over color channels~(HSV~color~space).
It can be assumed that field players are most frequently detected and that this is roughly balanced between both teams. Furthermore, due to the previous segmentation of the playing field, only a few detections are expected which are not field players. 
\emph{DBScan} requires two parameters:~$\epsilon$, which is the maximum normalized~(color) $L2$ distance between two detections to be assigned to the same cluster, and~$n_{cls} \in [0, 0.5]$, which specifies how many of all detections must belong together to form a cluster~(maximum of 0.5 due to two main clusters). 
Since the optimal value for $\epsilon$ will be different for each match, a grid search for randomly selected frames of each sequence of the match is performed to determine the parameter. 
In contrast to previous work that generally utilizes color histograms~\cite{lu2013learning, tong2011automatic} to reduce the input feature space, we apply the average over pixels 
without any performance decline.
The value for $\epsilon$ is selected, for which the cost function
    $c(\epsilon)=|X_{\text{(O)ther}}|+||X_{\text{A}}|-|X_{\text{B}}||$
is minimal and restricted to form exactly two clusters ($X_{\text{A}}$ and $X_{\text{B}}$). 
The cost function should ensure that the clusters A and B, which represent the two teams, are about the same size and that there are as few as possible unassigned detections ($X_{\text{(O)ther}}$).

\section{Experimental Results}
\label{sec:experiments}
All individual components are evaluated individually, while the main task of \emph{player position estimation} is evaluated at the end.
The main test sets that are used both to evaluate the sports field registration and \emph{player position estimation} are introduced in Section~\ref{exp:datasets}.
As shot boundary and shot type classification~(Section~\ref{sec:exp:shots}) are common pre-processing steps in video data, we refer to the supplemental material~\ref{apx:shot_boundary}. 
Section~\ref{exp:player_detection} and~\ref{exp:team_assignment} focus on the evaluation of player detection and team assignment, while
the evaluation of sports field registration is reported in Section~\ref{exp:h_estimation}.
Finally, the main task is evaluated by comparing the estimated positional data with the ground-truth data~(Section~\ref{exp:position_estimation}).

\subsection{Main Datasets}\label{exp:datasets}
To evaluate the main task of \emph{player position estimation}, synchronized video and positional data are needed.
To indicate the generalizability, we use a total of four datasets that are primarily designed only for testing, i.e., no training nor fine-tuning of individual modules is performed on this or closely related data.

Common broadcast videos of different resolutions~(SD and HD) and seasons~(2012, 2014) are available as well another type of 
video -- the tactic-cam~(TC): this camera recording is without any cuts and usually covers a wider range of the pitch.
Since the tactic-cam is located next to the main TV~camera and usually covers the majority of players, it is usually used for video analysis.
In general, each dataset contains four halves from four matches from the German Bundesliga in 25\,Hz temporal resolution with synchronized positional data.
Our datasets are referred to as \emph{TV12}~(2012, SD resolution), \emph{TV14}~(2014, HD), \emph{TC14}~(HD), and \emph{TV14-S} that covers the broadcast videos of the same matches as \emph{TC14}.  
Due to temporal inconsistencies in the raw video of \emph{TV14-S} to the positional data, these videos are synchronized using the visible game clock.
The position data are considered as ground truth since they are generated by a calibrated (multi-)camera system that covers the entire field. 
However, this system can be inaccurate in some cases~\cite{pettersen2018quantified}. 
An error of one meter is to be assumed in the data provided to us.

The quality of the field registration is essential for the accurate prediction of the player positions, but as there is only one limited dataset for sports field registration in soccer~\cite{homayounfar}, we manually estimate ground-truth homography matrices for a subset of our datasets.
In particular, 25 representative and challenging images per match are chosen to cover a wide range of camera settings resulting in 100 annotated images per test set.
The remaining modules, i.e., shot boundary and shot type classification, player detection, and team assignment are trained and evaluated on other publicly available datasets and introduced in their respective sections.

\subsection{Evaluating Player Detection}
\label{exp:player_detection}
Player detection and the usage of the homography matrix enable the extraction of two-dimensional coordinates for players. While a general object detector like \emph{Faster R-CNN}~\cite{fasterrcnn} localizes the bounding box for each object, this information is not necessarily needed, rather the exact position is of interest. 
To assess the performance of \emph{CenterTrack}~\cite{centertrack} on soccer data, 
we compare it to another specialized network~\cite{FootAndBall2020} for this domain and to a general object detection framework that is fine-tuned~\cite{fasterrcnn} for the soccer domain. 
We note that alternative solutions such as~\cite{light_cascaded} exist and a comparison is generally possible. 
However, it is out of scope of our paper to re-implement and test several variants especially if a satisfactory quality is achieved with the selected solution.

\textbf{Datasets \& Setup:}
Due to the lack of publicly available datasets for training and evaluation, \citet{FootAndBall2020} train their network on two small-scale datasets~\cite{d2009investigation, light_cascaded} where the training and test data is separated by frame-wise shuffling and subsequential random selection ($80\%$ training, $20\%$ test).
\emph{CenterTrack} can exploit temporal information to track players. 
However, to the best of our knowledge, there exists only one dataset in the domain of soccer with tracking information (\emph{ISSIA-CNR}~\cite{d2009investigation}), but it contains a very limited number of scene perspectives from multiple static cameras and is thus inappropriate for our system.
For a fair comparison with the alternative approach, we follow the train-test split of \citet{FootAndBall2020} where individual frames are used for training.
The publicly available \emph{ISSIA-CNR}~\cite{d2009investigation} dataset contains annotated sequences from several matches captured by six static cameras (in $30\,Hz$ and FHD resolution) comprising \num{3000} frames per camera. \emph{Soccer Player}~\cite{light_cascaded} is a dataset created from two professional matches where each match is recorded by three HD broadcast cameras with $30\,Hz$ and bounding boxes are annotated for approximately \num{2000} frames.
For evaluation, we report the average precision (AP) according to~\cite{AP_implementation}. In the final step of \emph{CenterTrack} bounding boxes are estimated, which makes AP a suitable metric to compare the performance of object detectors, even though the size of the bounding box is not relevant to extract positional data.
We refer to the supplemental material~(\ref{apx:impl_details}) for details about the training process.

\textbf{Results:} The results on the test set for our fine-tuned \emph{Faster R-CNN}~\cite{fasterrcnn}, \citet{FootAndBall2020}'s model and the fine-tuned \emph{CenterTrack}~\cite{centertrack} are reported in Table~\ref{tab:eval:detection}.
\begin{table}[tb!]
\setlength\extrarowheight{0pt}
\begin{center}
\small
\begin{tabularx}{\linewidth}{X|cc}
\toprule
 & ISSIA-CNR~\cite{d2009investigation} & Soccer Player~\cite{light_cascaded} \\ \midrule
Faster R-CNN~\cite{fasterrcnn} & 87.4 & \textbf{92.8} \\ 
FootAndBall~\cite{FootAndBall2020} & \textbf{92.1} & 88.5 \\ 
\textbf{CenterTrack}~\cite{centertrack} & 90.1 & 90.2 \\ 
\bottomrule
\end{tabularx}
\end{center}
\caption{Performance evaluation for player detection: The average precision in percent is measured on two subsets from the \emph{ISSIA-CNR} and \emph{Soccer Player} dataset.}
\label{tab:eval:detection}
\end{table}

Since \emph{Faster R-CNN} and \emph{FootAndBall} perform well on only one test set and perform significantly worse on the other, this suggests a lack of generalizability whereas \emph{CenterTrack} achieves good results on both data sets.
As \emph{CenterTrack} benefits from training with tracking data~\cite{centertrack}, we are confident that results can further be improved, but choose this model for our pipeline as it already provides good results.

\subsection{Evaluating Team Assignment}
\label{exp:team_assignment}

In this experiment, we evaluate the team assignment that relies on detected bounding boxes.

\textbf{Dataset \& Setup:}
In contrast to very small datasets~\cite{d2009investigation, light_cascaded}, \citet{dataset_shots}'s dataset provides a good diversity regarding the environmental setting~(camera movements, lighting conditions, different matches, jersey colors, etc.). Therefore, a subset from their dataset 
is manually annotated with respect to team assignment. 
To bypass errors in the player detection, bounding boxes and player assignment are manually annotated for a set of frames containing multiple shot perspectives and matches. Team assignment is annotated for three categories, \emph{team A}, \emph{team B}, and \emph{other} including referees and goalkeepers~(due to its sparsity). 
We randomly select one frame for a total of ten shots captured by the main camera for each match.
We took \num{20} matches that were already used to evaluate the temporal segmentation 
resulting in \num{200} frames for evaluation.
As mentioned before, the aim is to find two main clusters, and we found empirically that $n_{cls}=0.2$ provides good results for this task.
\newpage
\textbf{Metrics:} \citet{team_assign} proposed micro accuracy for this task, but this metric only considers labels from both teams and is insufficient in our case, since it can be misleading when the algorithm assigns uncertain associations to the class \emph{other}. To prevent this, referees and goalkeepers must be excluded from the object detection 
or an alternative metric needs to be defined.
For this reason, we additionally consider the macro accuracy for all three classes.

\textbf{Results:} Our simple method performs well, both in terms of macro accuracy~($0.91$) for the three classes and micro accuracy ($0.93$) for the two team classes. 
We found that most errors are players that are assigned to \emph{other} (goalkeepers, referees). This leads to the conclusion that field players are assigned correctly with a high probability in most cases.
In comparison to an end-to-end approach for team assignment of \citet{team_assign}, where the overall performance is evaluated on basketball data, a similar micro accuracy ($0.91$) is reported. However, the domain basketball differs much from soccer 
making a direct comparison difficult.

\subsection{Importance of Sports Field Registration}\label{exp:h_estimation}
As already introduced, many approaches rely on manually annotated ground-truth data for training. 
There exist only one public benchmark dataset (\textit{WorldCup2014}~(\textit{WC14})~\cite{homayounfar}). 
While the test set follows the same data distribution as the training data, in particular, the camera hyper-parameters~(location, focal-length, etc.), generalization capabilities are not investigated by existing solutions~\cite{nie2021robust, sha2020end, error_refine, chen2019}.
Primarily, to indicate the adaptability of \citet{chen2019}'s approach~(Section~\ref{sec:field_red}) to different environmental settings, we explore several hyper-parameters on our target test sets~(see Section~\ref{exp:datasets}). Additionally we compare them with recent work.

\textbf{Metrics:}
Since the visible part of the pitch is of interest for application, we report the intersection over union~($IoU_{\text{part}}$) score to measure the calibration accuracy.
It is computed between the two edge images using the predicted homography and the ground-truth homography on the visible part of the image.

\textbf{Camera Hyper-parameters:}
In general, we assume that the recommended parameters~\cite{chen2019}~(derived from \textit{WC14}~\cite{homayounfar}) for generating synthetic training data fit for many soccer stadiums. 
However, we also evaluate slight modifications of the base camera parameters which are available in \textit{WC14}: camera location distribution \small$\mathcal{N}(\mu=[52, -45, 17]^T, \sigma=[2, 9, 3]^T)$\normalsize~in meters, i.e., the average location from all stadiums~(origin is the lower left corner flag of the pitch); focal length~(\small$\mathcal{N}(3018, 716)\,mm$\normalsize) and pan~(\small$\mathcal{U}(-35^\circ, 35^\circ)$\normalsize), tilt~(\small$\mathcal{U}(-15^\circ, -5^\circ)$\normalsize) ranges. 
We extend the pan and tilt range to \small$(-40^\circ, 40^\circ)$\normalsize~and \small$(-20^\circ, -5^\circ)$\normalsize, respectively, in all models.
As the tactic-cam obviously covers a wider range (especially focal length as seen in Figure~\ref{fig:qualitative_results}~A,D,E), we also test versions, where we uniformly sample from the focal length parameters~(\small$\mathcal{U}_{\text{focal length}}(1000, 6000)$\normalsize)~ and from the locations~(\small$\mathcal{U}_{xyz}([45, -66, 10]^T, [60, -17, 23]^T)$\normalsize), and double the number of training images to \num{100000}.
Training process for line segmentation and homography estimation remain unchanged and we refer to \citet{chen2019} and \ref{apx:impl_details} for implementation details.

\begin{table}[tb]
\setlength{\tabcolsep}{1.8pt}
\setlength\extrarowheight{0pt}
\begin{center}
\small
\fontsize{8}{10}\selectfont
\def\arraystretch{0.9}
\begin{tabularx}{\linewidth}{X|rr|rrrrrr}
\toprule
\multicolumn{1}{c}{} & \multicolumn{2}{c}{WC14~\cite{homayounfar}} & \multicolumn{1}{c}{TV12} & \multicolumn{1}{c}{TV14} & \multicolumn{1}{c}{TC14} & \multicolumn{1}{c}{TV-S} \\
\multicolumn{1}{c}{Approach} & \multicolumn{1}{c}{Mean} & \multicolumn{1}{c}{Med.}  & \multicolumn{1}{c}{Mean (std)} & \multicolumn{1}{c}{Mean} & \multicolumn{1}{c}{Mean} & \multicolumn{1}{c}{Mean}\\ \midrule 
\cite{chen2019} repr. w.o. refinement                   & 88.3 & 90.2  & 63.6 (34.8) & 80.2 & 82.7 & 84.5 \\
\cite{chen2019} reproduced                              & 93.6 & 96.5  & 66.5 (37.4) & 85.2 & 88.7 & 90.2 \\
$\mathcal{U}_\text{focal len.}$                         & 92.2 & 96.6  & 64.3 (34.8) & 85.5 & \textbf{92.5} & 89.6 \\
$\mathcal{U}_\text{focal len.}$ + 2x num. cam.          & 94.6 & 96.2  & 59.5 (39.1) & 82.4 & 89.9 & \textbf{91.0} \\
$\mathcal{U}_\text{focal length}$ + $\mathcal{U}_{xyz}$ & 92.2 & 95.6  & 61.1 (38.3) & \textbf{87.4} & 87.1 & 89.8 \\
\citet{error_refine} (repr.$^{*}$)                      & 95.1 & 96.7 &  72.1 & 72.5 & 65.0 & 76.6 \\
CCBV~\cite{sha2020end}                                  & 94.2 & 95.4 & - & - &-  & -\\
\textcolor{gray}{Student CCBV~\cite{cioppa2021camera}$^\dag$} & 88.5 & 92.3 & -  & - & - & - \\ 
\textcolor{gray}{Teacher CCBV~\cite{sha2020end}~\cite{cioppa2021camera}} & 96.6 & 98.7 & - & - & - & -  \\ 
\textcolor{gray}{\citet{nie2021robust} keypoints}  & 95.8 & 97.2 & -  & - & - & - \\
\textcolor{gray}{\citet{nie2021robust} alignment} & 95.9 & 97.1 & -  & - & - & - \\
\bottomrule
\end{tabularx}
\end{center}
\caption{
Evaluation of multiple candidates for the sports field registration on several test sets using $IoU^{part}$.
$^{*}$official released model; $^\dag$no fine-tuning on WC14 but learned from private teacher model; Gray colored: private training data.
}
\label{tab:h_eval}
\end{table}

\textbf{Results:}
As reported in Table~\ref{tab:h_eval} the reproduced results~(base parameters) from \citet{chen2019} at \textit{WC14} are of similar quality compared to other methods~\cite{nie2021robust, sha2020end, error_refine}. 
We observe a noticeable drop in $IoU_{\text{part}}$ on our test sets where the camera parameters~(especially the camera position ($x,y,z$)) are unknown. 
For the \emph{TV12} test set all configurations fail on challenging images.
This further indicates that the original parameters are optimized for the camera dataset distribution in \textit{WC14}.
However, on the remaining three test sets, the approach of \citet{chen2019} is able to generalize
, whereas an alternative solution~\cite{jiang2020optimizing} fails.
Due to the non-availability of (private)~training data a comparison with~\cite{nie2021robust, sha2020end} is not fair~(colored gray). 
Yet, these approaches seem to yield comparable results.
A (student)~CCBV~\cite{sha2020end}~model from \cite{cioppa2021camera} is trained on the output of a teacher model.
As it was originally trained on a large-scale and private dataset, noticeable lower transfer performance is observed on \emph{WC14}. 
In summary, with slight changes in the hyper-parameters, the approach from \citet{chen2019} is suitable for the applicability to new data without fine-tuning by human annotations.

\subsection{Player Position Estimation}
\label{exp:position_estimation}
This section investigates the performance for player position estimation. 
Besides, errors of individual modules, i.e., sports field registration, player detection, and team assignment as well as compounding errors of the system are discussed.
We choose the full datasets as introduced in Section~\ref{exp:datasets}.
Despite the shot boundary and shot type classification provide good results~(see Appendix~\ref{apx:shot_boundary}), we eliminate their influence by considering manually annotated shots as the results for position estimation depend on this pre-processing step.
False-negative errors lead to a lower number of relevant frames for the system's output and for evaluation, while false-positive errors (e.g., close-ups) primarily produce erroneous output for homography estimation.

\paragraph{Metrics:} 
We measure the distance~(in meters) between the estimated positions and the actual positions by taking the mean and median over all individual frames~($d_\text{mean}$, $d_\text{med.}$) and additionally report how many frames have an error of less or equal than $l \in \{2.0, 3.0\}$ meters~($a_{l}$).
As previously mentioned, the sensor devices that capture position data~(used as ground-truth also in other works~\cite{memmert2017current}) can be slightly inaccurate. 
A domain expert confirmed, that errors in our system of less or equal than $l\leq 2\,m$ can be considered as correct results and that errors of less than $3\,m$ can still be meaningful for some sports analysis applications.

\begin{figure*}[tb!]
\begin{center}
\includegraphics[width=1.0\linewidth]{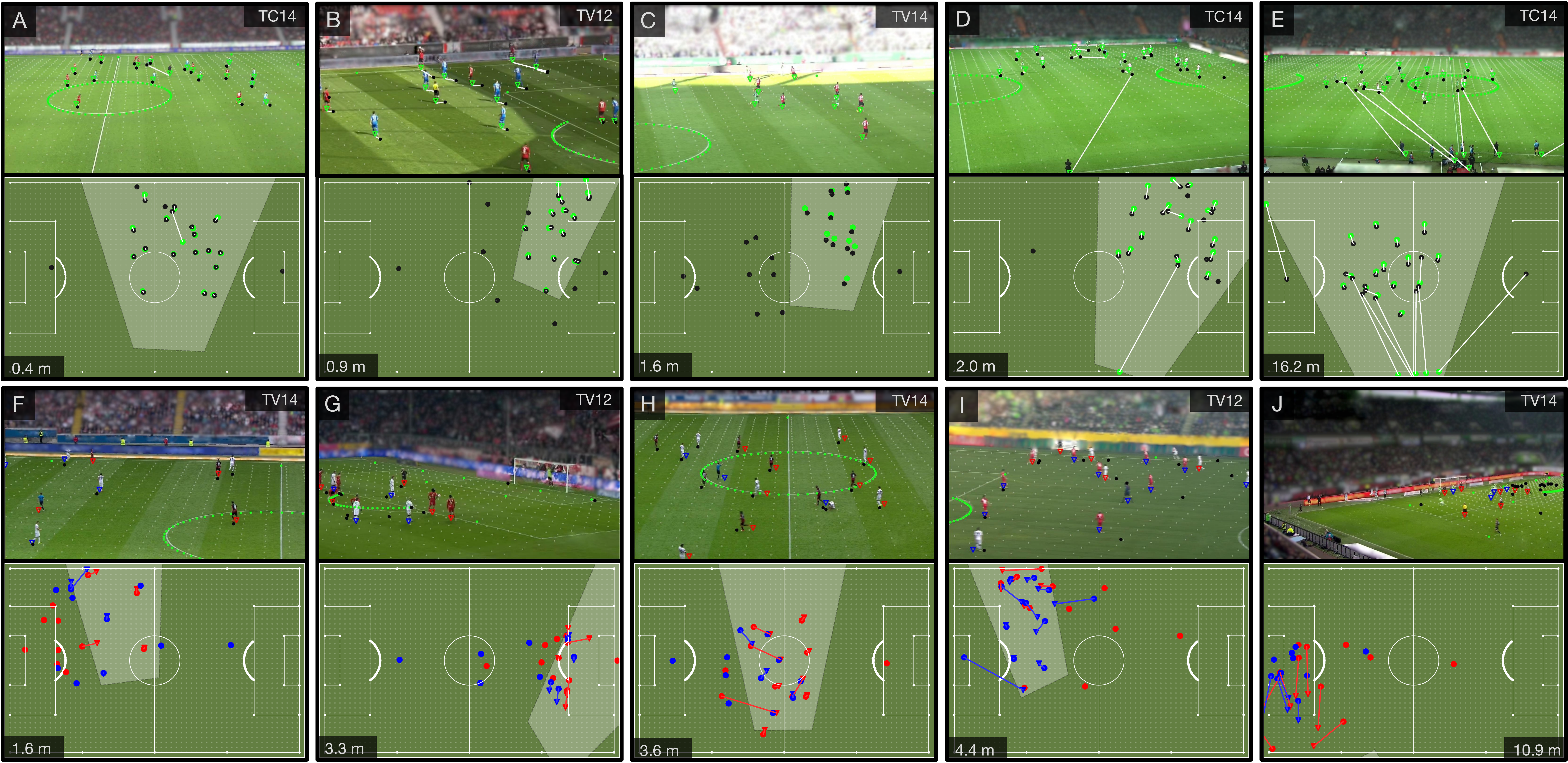}
\end{center}
\caption{Qualitative results of the proposed system for the extraction of positional player data ordered from low (left) to high error~(right): The top row presents the output without considering teams. The \textcolor{green}{green} triangles correspond to the predicted positions~($\bigtriangledown$) of players and the black points to the ground-truth positions~($\bullet$); team assignments are colorized \textcolor{red}{red} and \textcolor{blue}{blue}. 
For the input image the ground-truth positions are re-projected according to the estimated homography matrix; in the sports field some grid points are highlighted.}
\label{fig:qualitative_results}
\end{figure*}

\textit{Matching estimated positions to ground-truth:} Most of the time only a subset of players is visible in the broadcast videos and there is no information about which player is visible at a certain frame -- making evaluation complex.
As there is no direct mapping between predicted and ground-truth positions and the number of detections may vary, the resulting linear sum assignment problem first minimized using the \emph{Hungarian Method}~\cite{kuhn1955hungarian}. 
Its solution provides a set of distances for each field player visible in the frame $t$, formally $\mathbb{D}_t = \{d_1, \hdots, d_n\}$ where $n$ is the number of players and $d_i= \lVert \boldsymbol{\hat p} - \boldsymbol p \rVert_2$ is the distance between the estimated position~$\boldsymbol{\hat p} \in \mathbb{R}^2$ for the $i$-th player to its actual~(ground-truth) position $\boldsymbol p$. 
To aggregate the player distances of one frame, the use of the average distance as an error metric can be misleading as an outlier, e.g., a false-positive player detection (like a substitute player or goalkeeper) can be matched to a ground truth position with high distance~(Fig.~\ref{fig:qualitative_results}~\textit{D,E,H}).
These outliers can drastically affect the average distance and lead to wrong impressions.
To efficiently reject outliers without using an error threshold as an additional system parameter, we propose to report the average distances of the best $80$-percent position estimates.
Detailed results for this aggregation are included in the Appendix~\ref{apx:per-frame-agg}.

\textit{Player mismatch (pm) due to homography estimation \& player detection errors:} Despite the self-verification~(\emph{sv}) step~(Section~\ref{self-verification}) that discards erroneous homography estimations, we cannot directly evaluate whether the remaining homography matrices are correct, since some errors are not considered~(e.g., wrong focal length as in Fig.~\ref{fig:qualitative_results}~\textit{D}). 
To analyze the impact of very inaccurate homography estimations and major errors in player detection, we utilize ground-truth data to isolate these types of failures.  
We re-project all ground-truth positions to the image space according to the estimated homography matrix.
If the number of detected players differs significantly from the actual players then the homography is probably erroneous~(called player mismatch: \emph{pm}). We also define a tolerance range of 5\% of the image borders to include players that are at the boundary to avoid penalizing smallest discrepancies in the estimation of the homography matrix.
Finally, we discard all frames for evaluation that do not satisfy the following condition:
\begin{equation}
\label{eq:violation_number_of_players}
    \alpha_t := 1 - \zeta < \frac{|\mathbb{D}_t^{real}|}{|\mathbb{D}_t^{gt}|} < 1 + \zeta
\end{equation}
$\alpha_t$ is the indicator function whether a frame $t$ is discarded based on the ratio of detected players $|\mathbb{D}_t^{real}|$ and expected players $|\mathbb{D}_t^{gt}|$.
For example, assuming that ten
players are claimed to be visible, but only six are detected by the system, then we want to discard such discrepancies and set $\zeta=0.3$.
Furthermore, we incorporate a constraint to measure the results after team assignment by differentiating between the teams before the linear sum assignment 
and report the mean distance over both teams per frame.

\begin{table}[tb]
\setlength{\tabcolsep}{2.3pt}
\setlength\extrarowheight{0pt}
\begin{center}
\small
\fontsize{7}{10}\selectfont
\def\arraystretch{0.9}
\begin{tabularx}{\linewidth}{X|ccc|cccc|cccc}
\toprule
\multicolumn{1}{l}{} & \multicolumn{1}{l}{} & \multicolumn{1}{l}{}      & \multicolumn{1}{l}{} &                      \multicolumn{8}{c}{Team Assignment Constraint}                          \\
\multicolumn{3}{r}{}                                          & \multicolumn{1}{l}{}      & \multicolumn{4}{|c}{no}                              & \multicolumn{4}{|c}{yes}                               \\ \midrule
Dataset              & \emph{sv}            & \multicolumn{1}{c|}{\emph{pm}} & \multicolumn{1}{c}{Ratio} & \multicolumn{1}{|c}{$d_\text{mean}$} & \multicolumn{1}{c}{$d_\text{med.}$}
&\multicolumn{1}{c}{$acc_{2}$} &\multicolumn{1}{c}{$acc_{3}$} & \multicolumn{1}{c}{$d_\text{mean}$} & \multicolumn{1}{c}{$d_\text{med.}$} & \multicolumn{1}{c}{$acc_{2}$} & \multicolumn{1}{c}{$acc_{3}$} \\ \midrule
\multirow{3}{*}{\emph{TV12}} &                    & \multicolumn{1}{l|}{}       & 1.00   & 4.15  & 1.84 & 0.56  & 0.71    &  5.33  & 3.47 & 0.27 & 0.43 \\
                     & \checkmark           & \multicolumn{1}{l|}{}             & 0.90   & 2.52   & 1.66 & 0.61  & 0.77   &  4.06  & 3.31 & 0.30 & 0.46 \\
                     & \checkmark           & \multicolumn{1}{l|}{\checkmark}   & 0.79   & 2.10   & 1.55 & 0.64  & 0.80   &  3.39  & 2.99 & 0.34 & 0.51 \\ \hline
\multirow{3}{*}{\emph{TV14}} &                      & \multicolumn{1}{l|}{}     & 1.00   & 3.33 & 1.78  & 0.56  & 0.70   &  4.95  & 2.20  & 0.31 & 0.46 \\
                     & \checkmark           & \multicolumn{1}{l|}{}             & 0.84   & 2.82 & 1.71  & 0.55  & 0.71   &  3.57  & 1.81  & 0.34 & 0.51 \\
                     & \checkmark           & \multicolumn{1}{l|}{\checkmark}   & 0.72   & 2.29 & 1.64   & 0.60 & 0.77   &  3.17  & 1.71  & 0.35 & 0.53 \\ \hline
\multirow{3}{*}{\emph{TC14}}  &                      & \multicolumn{1}{l|}{}    & 1.00   & 3.71   & 1.20 & 0.74  & 0.83  &  3.32  & 1.39  & 0.65  & 0.78   \\
                     & \checkmark           & \multicolumn{1}{l|}{}             & 0.89   & 1.81   & 1.14 & 0.79  & 0.88   & 2.16  & 1.34   & 0.68 & 0.81  \\
                     & \checkmark           & \multicolumn{1}{l|}{\checkmark}   & 0.78   & 1.66   & 1.13 & 0.79  & 0.88   & 1.92  & 1.29   & 0.71 & 0.81 \\ \hline
\multirow{3}{*}{\emph{TV14-S}}  &                      & \multicolumn{1}{l|}{}  & 1.00   & 2.47 & 1.36   & 0.69  & 0.81  & 3.19   & 2.44 & 0.43 & 0.58   \\
                     & \checkmark           & \multicolumn{1}{l|}{}             & 0.92   & 1.89  & 1.29 & 0.73 & 0.85    & 2.89   & 2.34 & 0.44 & 0.59    \\
                     & \checkmark           & \multicolumn{1}{l|}{\checkmark}   & 0.75   & 1.73  & 1.27 & 0.75  & 0.87   & 2.78   & 2.32 & 0.45 & 0.60    \\
             
\bottomrule                
\end{tabularx}
\end{center}
\caption{Results regarding mean~($d_\text{mean}$) and median error~($d_\text{med}$) in meters and fraction of frames with an error of less or equal than $l$~meters~($acc_l$) of the total system on several datasets. \textit{Ratio} indicates how many frames are kept for evaluation after applying different criteria~(system output: only with \emph{sv}).}
\label{tab:pos_eval}
\end{table}

\paragraph{Results:} Table~\ref{tab:pos_eval} shows the results for each dataset while taking the best performing models from Table~\ref{tab:h_eval}. 
The results are summarized for all matches by taking the mean of the per match results.
The results after self-verification~\emph{sv} are the output of our system and set its tolerance area to $\rho=3\,m$. 
The results for other thresholds are reported in the supplemental material~(\ref{apx:apx:rho}). 
With the \emph{pm} criteria the impact of erroneous sports field registration is analyzed.
As evaluated, the applied model for sports field registration provides good results, however, for a couple of frames the $IoU_{\text{part}}$ is below $90\,\%$. 
Our \emph{sv}-process is able to discard some of these frames as the error drops significantly.
For the remaining frames, the pipeline provides promising results on all datasets. 
The \emph{pm} criterion demonstrates the high impact of the sports field registration.
Even for marginal errors in the homography estimation, i.e., ca. 95\% IoU, the absolute error in meter~(mean) is about 1\,m when back-projecting known keypoints~\cite{nie2021robust}. 
Hence, the applied reduction of bounding boxes to one point does not substantially affect the error in meters.
The qualitative examples in Figure~\ref{fig:qualitative_results}~(with applied \emph{sv}) primarily show the output of the pipeline and support the choice of our metrics. 
In Figure~\ref{fig:qualitative_results}~(\textit{I, J}), the output is obviously erroneous, but not discarded in the \emph{sv} process demonstrating the importance of an accurate sports field registration.
Furthermore, the influence of false-positive field players~(\textit{A,D,E}), and incorrect team identification is visible~(\textit{G,H,I}).
Since the quantitative results are weaker with team assignment, this suggests a lack of generalizability to the test data for the player detection and team assignment module.
Indeed, player tracking is not covered which would lead to more stable predictions across multiple frames, and temporal consistency of the sports field registration is not evaluated quantitatively. 
However, the sports field registration appears to provide stable results even without explicitly treating the temporal component but could be post-processed in an additional step~\cite{sharma, linnemann2013temporally}.

In summary, we claim that our system outputs promising results in many cases providing a first baseline to conduct various automatic analyses, for instance, regarding formation detection~\cite{bialkowski2014formation, ericjonas} or space control~\cite{fernandez2018wide, rein2017pass}.

\section{Conclusions \& Future Work}
\label{sec:conclusion}
In this paper, we have presented a fully-automated system for the extraction of positional data from broadcast soccer videos with interchangeable modules for shot boundary detection, shot type classification, player detection, field registration, and team assignment.
All components as well as their impact on the overall performance were evaluated. 
We investigated which parts of the pipeline influence each other and how they could be improved, e.g., by fine-tuning a specific module with more appropriate data. 
A relatively small error in meters should allow sports analysts to study team behavior. 
Indeed, the adaptation to other sports would definitely be interesting.
In the future, we also plan to integrate a tracking module.
However, additional steps for player re-identification~(within and across shots) are necessary to allow player-based analysis across a match.

\section*{Acknowledgement}

This project has received funding from the German Federal Ministry of Education and Research (BMBF~--~Bundesministerium für Bildung und Forschung) under~01IS20021B and~01IS20021A.

\section*{Appendix}
\appendix

In the supplemental material the shot boundary detection and shot type classification are evaluated in Section~\ref{apx:shot_boundary} and contains additional dataset description for the evaluation of the team assignment in Section~4.3 of the main paper. 
The computation costs to run the entire pipeline is roughly specified in Section~\ref{apx:runtime} whereas
related limitations regarding the temporal consistency is addressed in Section~\ref{apx:temporal_consistency}.
Section~\ref{apx:rho} provides additional information to the self-verification filtering to discard obviously errors in the sports field registration, especially the influence of the controllable parameter~$\rho$.
The choice of the function to aggregate per-frame distances~(predicted positions matched to ground-truth) is described in Section~\ref{apx:per-frame-agg}.

\section{Shot Boundary \& Shot Type Detection}\label{apx:shot_boundary}
In this experiment, we evaluate the performance of the pre-processing module on soccer data that includes (1)~shot boundary detection and (2)~shot type detection where its output is essential for the success of the following modules.
\paragraph{Dataset:}
For this purpose, we select a subset from \emph{SSET}~\cite{dataset_shots} where the shot boundaries and camera types (main camera or other) are manually annotated for each frame. Since a general method for shot boundary detection is applied where no extra training is needed, 20 matches are randomly selected that contain different challenges like varying resolutions (ten HD ($1280\times720$) and ten SD ($640\times360$)), light conditions (artificial vs. daylight) or camera movement.

\paragraph{Shot Boundary Detection:} To evaluate the shot boundary detection, we report recall, precision, and the $F_1$ score for pre-trained\footnote{\scriptsize \url{https://github.com/soCzech/TransNetV2}} models of \emph{TransNet}~\cite{transnet} and \emph{TransNetV2}~\cite{transnetv2}.
However, minor temporal offsets of detected cuts~(e.g., a cut was detected a few frames earlier or later than the ground truth) are treated as missing cuts.
Therefore we consider three different tolerance offsets with $\Delta=\{0, 1, 2\}$ frames.
The results for \emph{TransNet} on soccer data are presented in Table~\ref{tab:shot_boundary_detection}.
With a tolerance of only two frames, the improved \emph{TransNetV2}~\cite{transnetv2} achieves useful results ($F_1=0.89$). Because this model is a general detector, fine-tuning is a way to improve the results. But this would be beyond the scope of this work and we believe that the performance is already sufficient.

\begin{table}[tbp]
\setlength{\tabcolsep}{2.3pt}
\setlength\extrarowheight{0pt}
\begin{center}
\small
\fontsize{7}{10}\selectfont
\def\arraystretch{0.9}
\begin{tabularx}{\linewidth}{X|lll|lll}
\toprule
 & \multicolumn{3}{c}{TransNet~\cite{transnet}} & \multicolumn{3}{c}{TransNetV2~\cite{transnetv2}} \\ \hline
$\Delta$ & \multicolumn{1}{c}{0} & \multicolumn{1}{c}{1} & \multicolumn{1}{c}{2} & \multicolumn{1}{c}{0} & \multicolumn{1}{c}{1} & \multicolumn{1}{c}{2} \\ \hline
$precision$ & 0.65 & 0.81 & \textbf{0.88} & 0.59 & 0.76 & 0.86 \\
$recall$ & 0.80 & 0.83 & 0.84 & 0.88 & 0.91 & \textbf{0.92} \\
$F_1$ & 0.72 & 0.82 & 0.86 & 0.71 & 0.83 & \textbf{0.89} \\
\bottomrule
\end{tabularx}
\end{center}
\caption{Evaluation of the shot boundary detection with $\Delta$ frames tolerance for the two methods.}
\label{tab:shot_boundary_detection}
\end{table}

\paragraph{Shot Type Detection:} Please recall, that it is reasonable to consider only shots captured by the main camera since it shows most of the players on the field. 
To obtain these shots, a simple but effective classification method was introduced in Section~3.1 of the main paper where only a pre-defined parameter~$\tau$ and the model for sports field registration are involved. 
We have determined this value experimentally using five randomly sampled matches from \emph{SSET}~\cite{dataset_shots} without intersections to the test set und using the reproduced model from \citet{chen2019}~(Table 2 second row of the main paper). For each match, a grid search is conducted to determine a corresponding threshold and the average overall matches provide a final value of $\tau=0.35$. The $F_1$ score is optimized as it provides the harmonic mean between precision and recall.
Applied on the test set, an overall $F_1$ score of $0.88$ indicates that almost all relevant shots are correctly classified with a precision of $0.90$ and recall of $0.87$.

\section{Runtime of the Pipeline}\label{apx:runtime}
All components are real-time capable ($>25\,fps$) on common hardware (we used a 8x 2.9 GHz CPU and one Nvidia 2080Ti to run each module) except for the field registration, especially the homography refinement~\cite{baker2004lucas}\footnote{\scriptsize \url{https://docs.opencv.org/4.5.1/dc/d6b/group\_\_video\_\_track.html\#ga1aa357007eaec11e9ed03500ecbcbe47}}. 
Even an alternative approximated solution~\cite{jiang2020optimizing} is not able to improve the runtime. However, this step is fundamental for the quality of the field registration. 
If computation time is a requirement, the entire module could be replaced for instance by \citet{nie2021robust} that was recently published, but unfortunately, the authors do not provide data, nor pre-trained models. 

\section{Temporal Consistency}\label{apx:temporal_consistency}
In the last two experiments~(Section~4.4 and~4.5 of the paper) we evaluated only single frames and did not consider temporal consistency of successive frames.
A previously evaluated approach~\cite{sharma} that detects individual outliers in the homography estimation and makes the estimates consistent across multiple frames~(a kind of smoothing) was not implemented for two reasons.
On the one hand, the runtime would have dropped to less than $1\,fps$ and on the other hand the authors have already shown that there are only minor improvements. 

\section{Self-Verification~(\emph{sv}) Parameter~$\rho$}\label{apx:rho}

In Section~3.3 of the paper we have introduced the self-verification~(\emph{sv}) criterion that allows the system to automatically  discard  individual  frames  where  the  sports field registration is erroneous.  
Here, we vary the control-able parameter $\rho$, and show its influence on the \emph{TV12} dataset in Table~\ref{tab:rho} according to the final result table~(Table~3).
The stricter (smaller) the value is chosen, the more frames are discarded without much improvement of the results. Thus, we have used $\rho=3\,m$ as it provides a good trade-off.

\begin{table}[]
\setlength{\tabcolsep}{2.3pt}
\setlength\extrarowheight{0pt}
\begin{center}
\small
\fontsize{7}{10}\selectfont
\def\arraystretch{0.9}
\begin{tabularx}{\linewidth}{X||c|c|cccc}
\toprule
$\rho$ & Ratio & $d_\text{mean}$ & $d_{\leq 2\,m}$ & $d_{\leq 3\,m}$ & $d_{\leq 4\,m}$ & $d_{\leq 5\,m}$   \\ \hline
1.0    &  0.81     & 2.12     &  0.70     &  0.84     &  0.91     &   0.94      \\
2.0    &  0.87       & 2.12     & 0.70      & 0.84      & 0.91      & 0.94        \\
3.0    &  0.90       & 2.15     & 0.69      & 0.84      & 0.90      & 0.93        \\
4.0    &  0.91     &  2.18    & 0.69      &  0.83     &  0.90     & 0.93        \\
5.0    &  0.92     &  2.20    & 0.68      &  0.83     &  0.90     & 0.93   \\
$\infty$    &  1.0     &  3.69    & 0.64      &  0.78     &  0.84     & 0.88   \\
\bottomrule
\end{tabularx}
\end{center}
\caption{Results of the actual error made in meters~($d_\text{mean}$) and fraction of frames smaller than $l$ meters on \emph{TV12} while varying the self-verification parameter $\rho$. \textit{Ratio} indicates how many frames are discarded for evaluation~($\infty$ corresponds to no self-verification.)}
\label{tab:rho}
\end{table}

\section{Per-Frame Distances Aggregation}\label{apx:per-frame-agg}
As shortly explained in the main paper, a set of per-frame distances~$\mathbb{D}_t = \{d_1, \hdots, d_n\}$ is aggregated to calculate a global error in meters over all frames and matches.
The use of the average distance as error metric can be misleading as an outlier, i.e., a false positive player detection~(e.g. substitute player) can be matched to a ground truth player position with high distance. These outliers can drastically affect the average distance and lead to wrong impressions.
%
To efficiently reject outliers without using an error threshold as additional system parameter, we proposed to report the average distances of the best $q$-percent position estimates.
The choice of $q$ is intuitively chosen with 80\% to allow minor false-positive player detection errors.

\begin{figure}[tb]
\begin{center}
\includegraphics[width=0.75\linewidth]{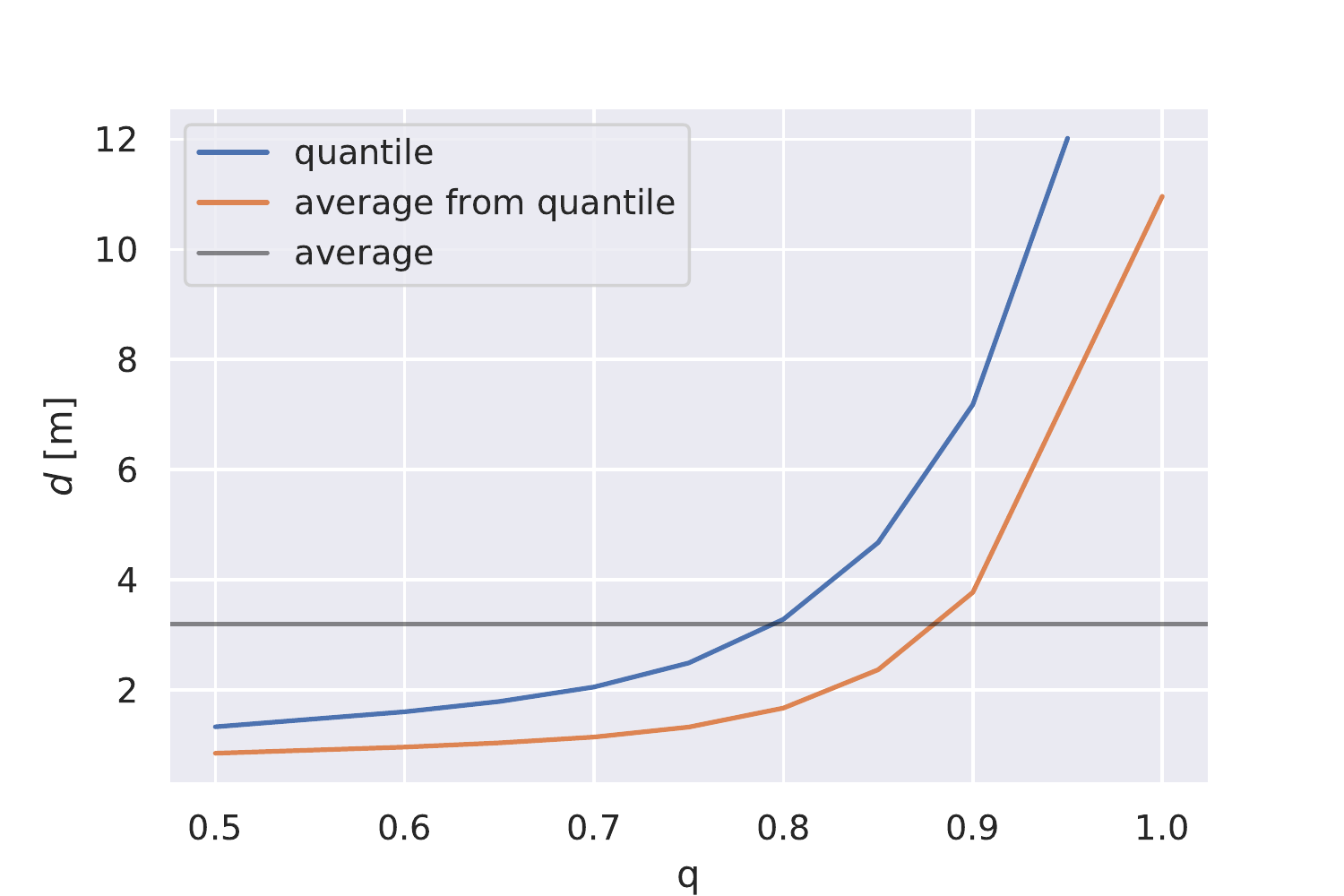}
\end{center}
   \caption{Error in meter over all frames for one match by varying $q$ when the quantile is used to aggregate per-frame distances: Ground-truth positions of the referees are not included which results in a lot of false-positive player detections.}
\label{fig:quantile_missing_gt_referees}
\end{figure}

\begin{figure}[bt]
\begin{center}
\includegraphics[width=0.75\linewidth]{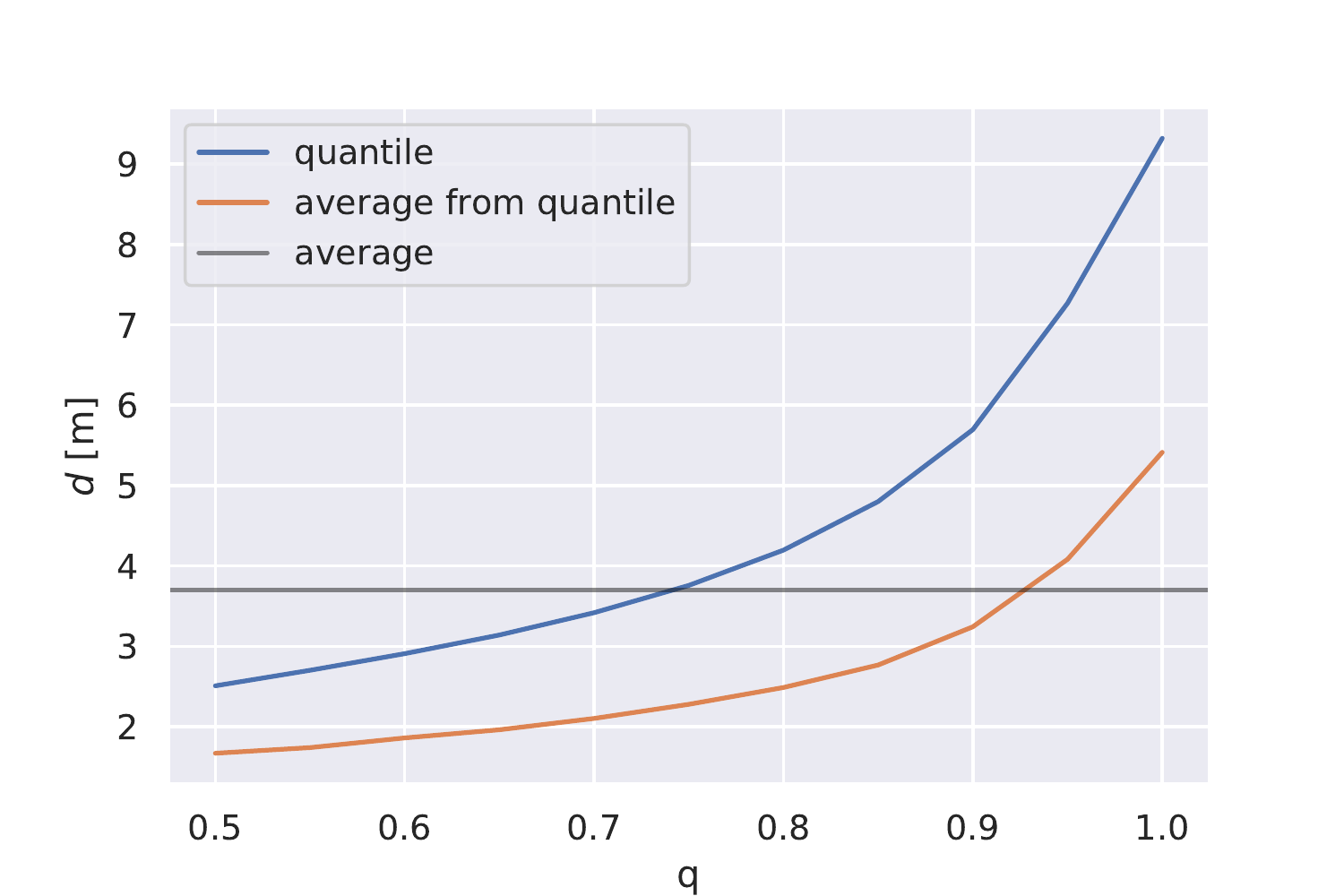}
\end{center}
   \caption{Error in meter over all frames for one match by varying $q$ when the quantile is used to aggregate per-frame distances.}
\label{fig:quantile}
\end{figure}

The influence of $q$ to the global error in meter over all frames~($d_\text{mean}$) of one randomly chosen match is addressed in Figure~\ref{fig:quantile_missing_gt_referees} and Figure~\ref{fig:quantile} after applying both filtering criteria~($sv$ and $pm$) and without considering team assignment. This influence of $q$ could be ideally explained for the match in Figure~\ref{fig:quantile_missing_gt_referees} where no ground-truth positions for the referees are given, but for most of time one referee is visible in the image and has a bounding box according to \emph{CenterTrack}~\cite{centertrack}. This bounding will be assigned to one of the ground-truth player positions which is just an issue in evaluation metric.

If we choose a $q$ between 0.8 and 0.9, these kinds of errors can be ignored in the evaluation because the focus is on the distances of the significant player positions.
A value of $q=0.8$ is also according to Figure~\ref{fig:quantile} not too strict.
The influence on the global error in meter using \emph{mean, median} and our \emph{proposed}  per-frame aggregation function is shown for \emph{TV12} without team assignment in Table~\ref{tab:apx:pos_eval}.

\begin{table}[tbhp]
\setlength{\tabcolsep}{2.3pt}
\setlength\extrarowheight{0pt}
\begin{center}
\small
\fontsize{7}{10}\selectfont
\def\arraystretch{0.9}
\begin{tabularx}{\linewidth}{X|ccc|ccc}
\toprule
Aggregation              & \emph{sv}            & \emph{pm} & Ratio & $d_\text{mean}$ & $d_{\leq 2\,m}$ &$d_{\leq 3\,m}$ \\ \hline
\multirow{3}{*}{Mean} &                    & \multicolumn{1}{l|}{}             & 1.00   & 8.90  & 0.48 & 0.69  \\
                     & \checkmark           & \multicolumn{1}{l|}{}             & 0.90   & 2.67   & 0.53 & 0.75    \\
                     & \checkmark           & \multicolumn{1}{l|}{\checkmark}   & 0.75   & 2.27   & 0.56 & 0.79   \\ \hline
\multirow{3}{*}{Median} &                    & \multicolumn{1}{l|}{}             & 1.00   & 3.70  & 0.62 & 0.78  \\
                     & \checkmark           & \multicolumn{1}{l|}{}             & 0.90   & 2.24   & 0.66 & 0.83    \\
                     & \checkmark           & \multicolumn{1}{l|}{\checkmark}   & 0.75   & 1.83   & 0.71 & 0.88   \\ \hline
\multirow{3}{*}{proposed} &                    & \multicolumn{1}{l|}{}             & 1.00   & 3.69  & 0.64 & 0.78  \\
                     & \checkmark           & \multicolumn{1}{l|}{}             & 0.90   & 2.15   & 0.69 & 0.84    \\
                     & \checkmark           & \multicolumn{1}{l|}{\checkmark}   & 0.75   & 1.77   & 0.73 & 0.88   \\
\bottomrule                
\end{tabularx}
\end{center}
\caption{The influence of the per-frame aggregation function on the actual error made in meters is shown on the \emph{TV12} dataset.}
\label{tab:apx:pos_eval}
\end{table}

\section{Implementation Details}\label{apx:impl_details}
\subsection{Sports Field Registration}
For sports field registration and line segmentation we have modified the officially available source code\footnote{\scriptsize \url{https://github.com/lood339/SCCvSD}}~\cite{chen2019}. 

During training, we sample 50000 camera poses using the provided camera pose engine. 
Camera parameters are set as explained in our paper. 
Following~\cite{chen2019} we generate edges images ($1280\times720$ resolution) for
sampled camera poses. Then, the edge images are resized
to $320\times 180$ and are used to train the two \emph{pix2pix} models with underlying \emph{U-Net} architecture.
We refer to the implementation section of~\cite{chen2019} for hyper-parameters as these parameters have remained unchanged.  
These generated edge images are the input of the siamese network~(structure as in \cite{chen2019}) and produces a 16-dimensional deep feature for each edge images used for retrieval during inference.

\subsection{Player Detection}
As a pre-trained \emph{CenterTrack}~\cite{centertrack} model is fine-tuned on the described data as in the paper, we use the fine-tuning script from the official implementation\footnote{\scriptsize \url{https://github.com/xingyizhou/CenterTrack}} with default parameters with the except that the number of epochs is set to 70.
In this context, we would like to point out that providing all details and hyperparameters in this document is not purposeful.

{\small
\setlength{\bibsep}{0pt} 
\bibliographystyle{ieee_fullname} 
\bibliography{egbib}
}

\end{document}